\documentclass[10pt,twocolumn,letterpaper]{article}

\usepackage{iccv}
\usepackage{times}
\usepackage{epsfig}
\usepackage{graphicx}
\usepackage{amsmath}
\usepackage{amssymb}
\usepackage{rotating}
\usepackage{array}
\usepackage{microtype}
\usepackage[shortcuts]{extdash}

\usepackage[pagebackref=true,breaklinks=true,letterpaper=true,colorlinks,bookmarks=false]{hyperref}

\iccvfinalcopy

\ificcvfinal\fi
\begin{document}

\title{Perceptually Motivated Method for Image Inpainting Comparison}

\author{Ivan Molodetskikh, Mikhail Erofeev, Dmitry Vatolin\\
Lomonosov Moscow State University\\
Moscow, Russia\\
{\tt\small ivan.molodetskikh@graphics.cs.msu.ru}
}

\maketitle

\begin{abstract}
   The field of automatic image inpainting has progressed rapidly in recent
   years, but no one has yet proposed a standard method of evaluating
   algorithms. This absence is due to the problem's challenging nature:
   image-inpainting algorithms strive for realism in the resulting images, but
   realism is a subjective concept intrinsic to human perception. Existing
   objective image-quality metrics provide a poor approximation of what humans
   consider more or less realistic.

   To improve the situation and to better organize both prior and future
   research in this field, we conducted a subjective comparison of nine
   state-of-the-art inpainting algorithms and propose objective quality metrics
   that exhibit high correlation with the results of our comparison.
\end{abstract}

\begin{figure*}
\begin{center}
   \includegraphics[width=\linewidth]{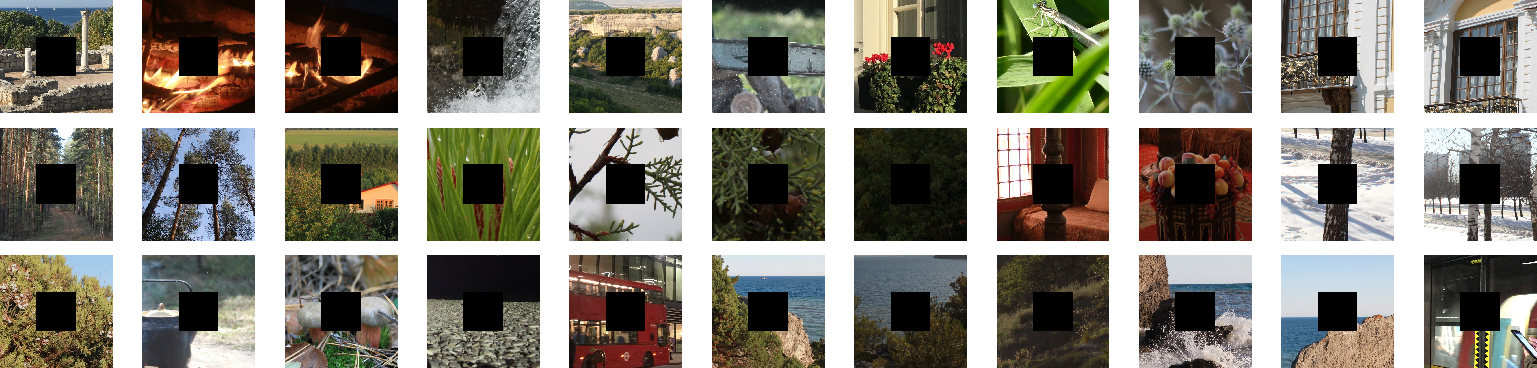}
\end{center}
   \caption{Images for the subjective inpainting comparison. The black square
            in the center is the area to be inpainted.}
   \label{fig:subjective-comparison-photos}
\end{figure*}

\section{Introduction}

Image inpainting, or hole filling, is the task of filling in missing parts of
an image. Given an incomplete image and a hole mask, an inpainting algorithm
must generate the missing parts so that the result looks realistic. Inpainting
is a widely researched topic. Many classical algorithms have been
proposed~\cite{telea2004image,criminisi2004region}, but over the past few years
most research has focused on using deep neural networks to solve this
problem~\cite{Pathak_2016_CVPR,song2017image,liu2017semantically,li2018context,Yu_2018_CVPR,Liu_2018_ECCV,yu2018free}.

Naturally, because of the many avenues of research in this field, the need to
evaluate algorithms emerges. The specifics of image inpainting mean this
problem has no simple solution. The goal of an inpainting algorithm is to make
the final image as realistic as possible, but image realism is a concept
intrinsic to humans.  Therefore, the most accurate way to evaluate an
algorithm’s performance is a subjective experiment where many participants
compare the outcomes of different algorithms and choose the one they consider
the most realistic.

Unfortunately, conducting a subjective experiment involves considerable time
and resources, so many authors resort to evaluating their proposed methods
using traditional objective image-similarity metrics such as PSNR, SSIM and
mean $l_2$ loss relative to the ground-truth image. This strategy, however, is
inadequate. One reason is that evaluation by measuring similarity to the
ground-truth image assumes that only a single, best inpainting result
exists---a false assumption in most cases. As a trivial example, consider that
inpainting is frequently used to erase unwanted objects from photographs.
Therefore, at least two realistic outcomes are possible: the original
photograph with the object and the desired photograph without the object.
Furthermore, we show that the popular SSIM image-quality estimation metric
correlates poorly with human responses and that the inpainting result can seem
more realistic to a human than the original image does, limiting the
applicability of full-reference metrics to this task.

Moreover, owing to the lack of a clear and objective way to evaluate inpainting
algorithms, different authors present results of different metrics on different
data sets when considering their proposed algorithms. Comparing these
algorithms is therefore even harder.

Thus, a perceptually motivated objective metric for inpainting-quality
assessment is desirable. The objective metric should approximate the notion of
image realism and yield results similar to those of a subjective study when
comparing outputs from different algorithms.

We conducted a subjective evaluation of nine state-of-the-art classical and
deep-learning-based approaches to image inpainting. Using the results, we
examine different methods of objective inpainting-quality evaluation, including
both full-reference methods (taking both the resulting image and the
ground-truth image as an input) and no-reference methods (taking just the
resulting image as an input).

It is important to note that we are \textit{not} proposing objective
quality-evaluation models trained on a database of subjective scores, as
obtaining sufficiently large and diverse databases of this nature is
impractical. We do, however, evaluate the proposed models by comparing their
correlations to human responses.

\section{Related work}

Little work has been done on objective image inpainting-quality evaluation or
on inpainting detection in general. The somewhat related field of
manipulated-image detection has seen moderate research, including both
classical and deep-learning-based approaches. This field focuses on detecting
altered image regions, usually involving a set of common manipulations:
copy-move (copying an image fragment and pasting it elsewhere in the same
image), splicing (pasting a fragment from another image), fragment removal
(deleting an image fragment and then performing either a copy-move or
inpainting to fill in the missing area), various effects such as Gaussian blur
and median filtering, and recompression (usually indicating that the image was
handled in a photo editor). Among these manipulations, the most interesting for
this work is fragment removal with inpainting.

The classical approaches to image-manipulation detection
include~\cite{pun2015image,li2017image}, and the deep-learning-based approaches
include~\cite{Bappy_2017_ICCV,zhu2018deep,salloum2018image,Zhou_2018_CVPR}.
These algorithms aim to locate the manipulated image regions by outputting a
mask or a set of bounding boxes enclosing suspicious regions. Unfortunately,
they are not directly applicable to inpainting-quality estimation because they
have a different goal: whereas an objective quality-estimation metric should
strive to accurately compare realistically inpainted images similar to the
originals, a forgery-detection algorithm should strive to accurately tell one
apart from the other.

\section{Inpainting subjective evaluation}%
\label{sec:subjective-evaluation}

The gold standard for evaluating image-inpainting algorithms is human
perception, since each algorithm strives to produce images that look the most
realistic to humans. Thus, to obtain a baseline for creating an objective
inpainting-quality metric, we conducted a subjective evaluation of multiple
state-of-the-art algorithms, including both classical and deep-learning-based
ones. To assess the overall quality and applicability of the current approaches
and to see how they compare with manual photo editing, we also included
human-produced images. We asked several professional photo editors to fill in
missing regions of the test photos just like an automatic algorithm would.

\subsection{Test data set}

Since human photo editors were to perform inpainting, our data set excluded
publicly available images: we wanted to ensure that finding the original photos
online and achieving perfect results would be impossible. We therefore created
our own private set of test images by taking photographs of various outdoor
scenes, which are the most likely target for inpainting.

Each test image was $512\times512$ pixels with a square hole in the middle
measuring $180\times180$ pixels. We chose a square instead of a free-form shape
because one algorithm in our comparison~\cite{Yang_2017_CVPR} lacks the ability
to fill in free-form holes. The data set comprised 33 images in total.
Figure~\ref{fig:subjective-comparison-photos} shows examples.

\subsection{Inpainting methods}

\begin{figure}[t]
\begin{center}
   \setlength{\tabcolsep}{1pt}
   \begin{tabular}{ m{2ex} m{0.3\linewidth} m{0.3\linewidth} m{0.3\linewidth} }
      & \multicolumn{1}{c}{Artist \#1}
      & \multicolumn{1}{c}{Artist \#2}
      & \multicolumn{1}{c}{Artist \#3} \\
      \rotatebox{90}{Urban Flowers}
      & \includegraphics[width=\linewidth]{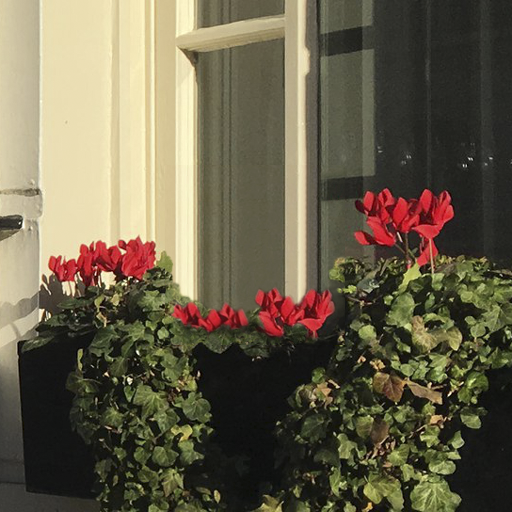}
      & \includegraphics[width=\linewidth]{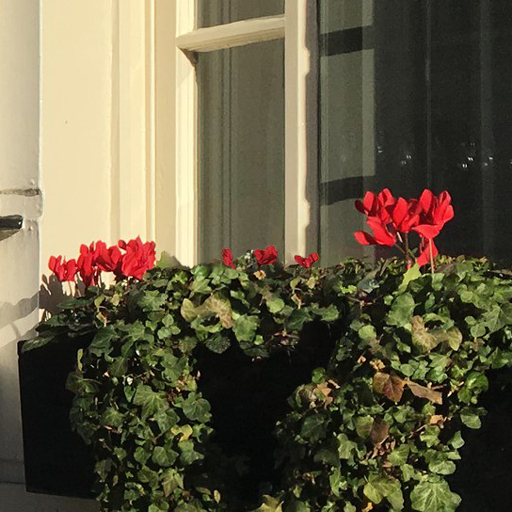}
      & \includegraphics[width=\linewidth]{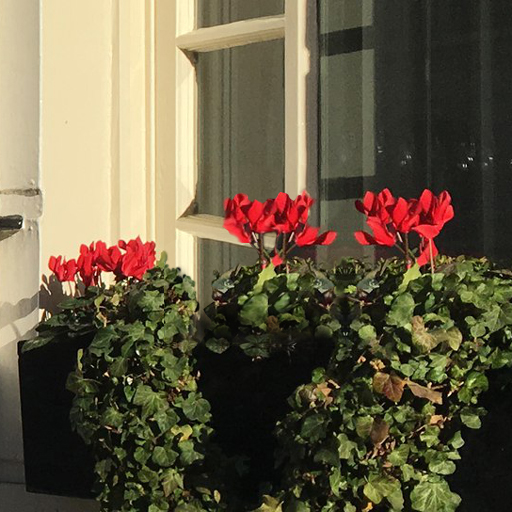} \\
      \rotatebox{90}{Splashing Sea}
      & \includegraphics[width=\linewidth]{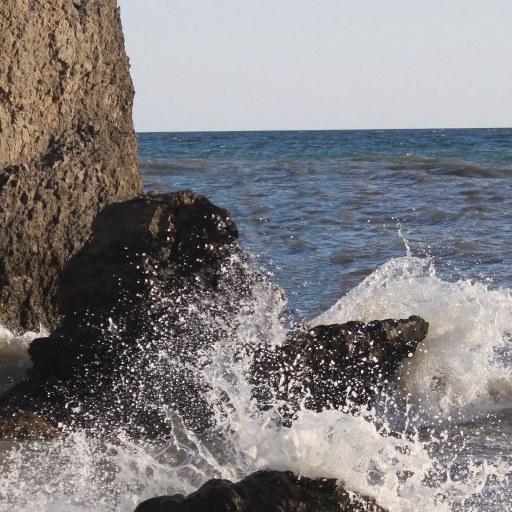}
      & \includegraphics[width=\linewidth]{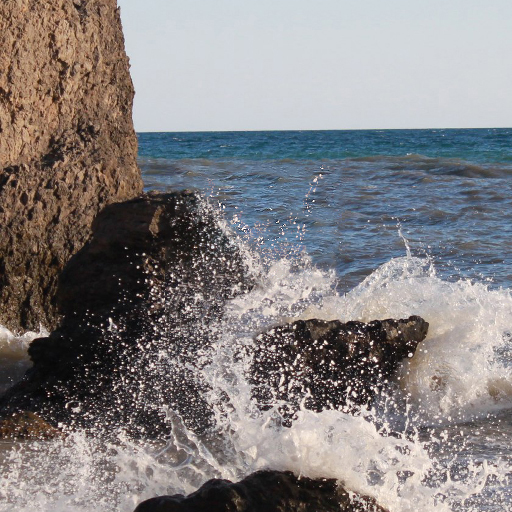}
      & \includegraphics[width=\linewidth]{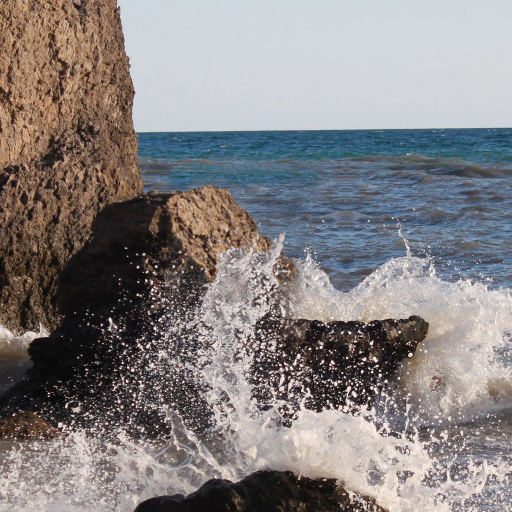} \\
      \rotatebox{90}{Forest Trail}
      & \includegraphics[width=\linewidth]{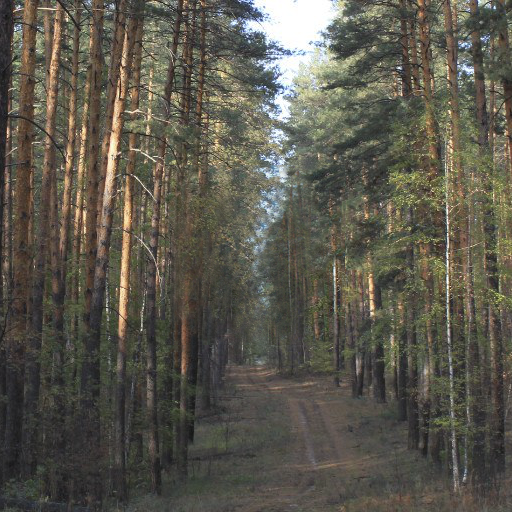}
      & \includegraphics[width=\linewidth]{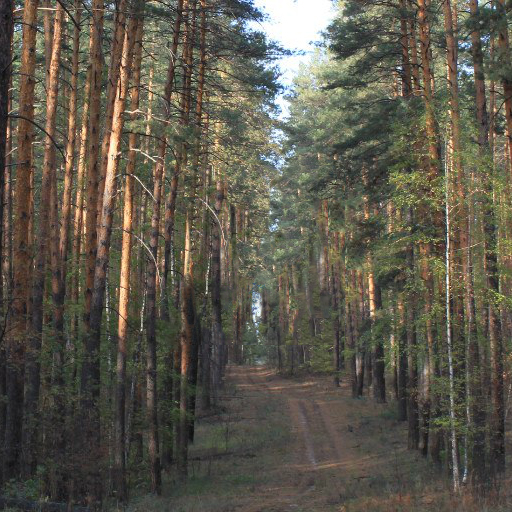}
      & \includegraphics[width=\linewidth]{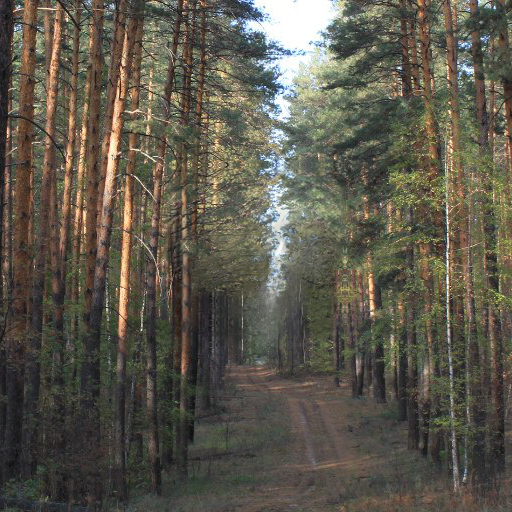} \\
   \end{tabular}
\end{center}
   \caption{Three images from our test set, inpainted by three human artists.}
   \label{fig:artist-results}
\end{figure}

We evaluated nine classical and deep-learning-based approaches:
\begin{itemize}
   \item Exemplar-based inpainting~\cite{criminisi2004region} is a well-known
      classical algorithm that finds the image patches that are most similar to
      the region being filled and copies them into that region in a particular
      order to correctly preserve image structures.

   \item Statistics of patch offsets~\cite{he2012statistics} is a more recent
      classical algorithm that employs distributions of statistics for the
      relative positions of similar image patches.

   \item Content-aware fill is a tool in the popular photo-editing software
      Adobe Photoshop. We picked it because, being part of a popular image
      editor, it is highly likely to be used for inpainting. Thus, comparing it
      with other state-of-the-art approaches is valuable. We tested the
      content-aware fill in Adobe Photoshop CS5; its implementation preceded
      deep learning’s explosion in popularity.

   \item Deep image prior~\cite{Ulyanov_2018_CVPR} is an unconventional
      deep-learning-based method. It relies on deep generator networks
      converging to a realistic image from a random initial state rather than
      by learning realistic image priors from a large training set.

   \item Globally and locally consistent image
      completion~\cite{iizuka2017globally} is a deep-learning-based approach
      that uses global and local discriminators, thereby improving both the
      coherence of the resulting image as a whole and the local consistencies
      of image patches.

   \item High-resolution image inpainting~\cite{Yang_2017_CVPR}, another
      deep-learning-based method, employs multiscale neural patch synthesis to
      preserve both contextual structures and high-frequency details in
      high-resolution images.

   \item Shift-Net~\cite{Yan_2018_ECCV} is a U-Net architecture that implements
      a special shift-connection layer to improve the inpainting quality. The
      shift-connection layer offsets encoder features of known regions to
      estimate the missing ones.

   \item Generative image inpainting with contextual
      attention~\cite{Yu_2018_CVPR} uses a two-part convolutional neural
      network to first predict the structural information and then restore the
      fine details. It also has a special contextual-attention layer that finds
      the most similar patches from the known image areas to aid generation of
      fine details.

   \item Image inpainting for irregular holes using partial
      convolutions~\cite{Liu_2018_ECCV} follows the idea that regular
      convolutional layers in deep inpainting networks are suboptimal because
      they unconditionally use both valid (known) and invalid (unknown) pixels
      from the input image. This method implements a partial convolution layer
      that masks out the unknown pixels.
\end{itemize}

Additionally, we hired three professional photo\-/restoration and
photo-retouching artists to manually inpaint three randomly selected images
from our test data set. To encourage them to produce the best possible results,
we offered a 50\% honorarium bonus for the ones that outranked the competitors.
Although we imposed no strict time limit, all three artists completed their
work within 90 minutes. Figure~\ref{fig:artist-results} shows their results.

\begin{figure}[t]
\begin{center}
   \includegraphics[width=\linewidth]{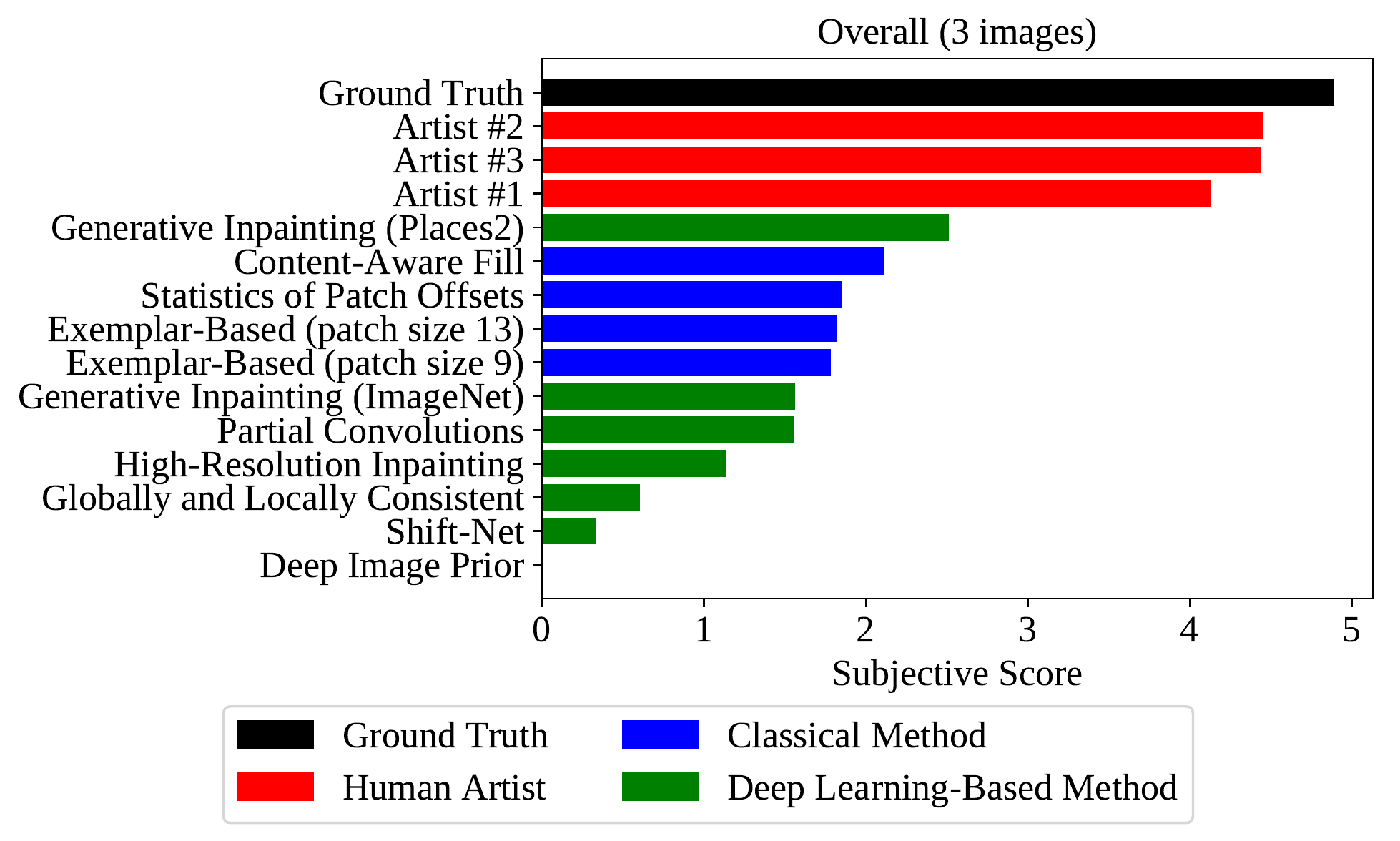}
\end{center}
   \caption{Subjective-comparison results across three images inpainted by
            human artists.}
   \label{fig:human-results-overall}
\end{figure}

\begin{figure*}
\begin{center}
   \includegraphics[width=\linewidth]{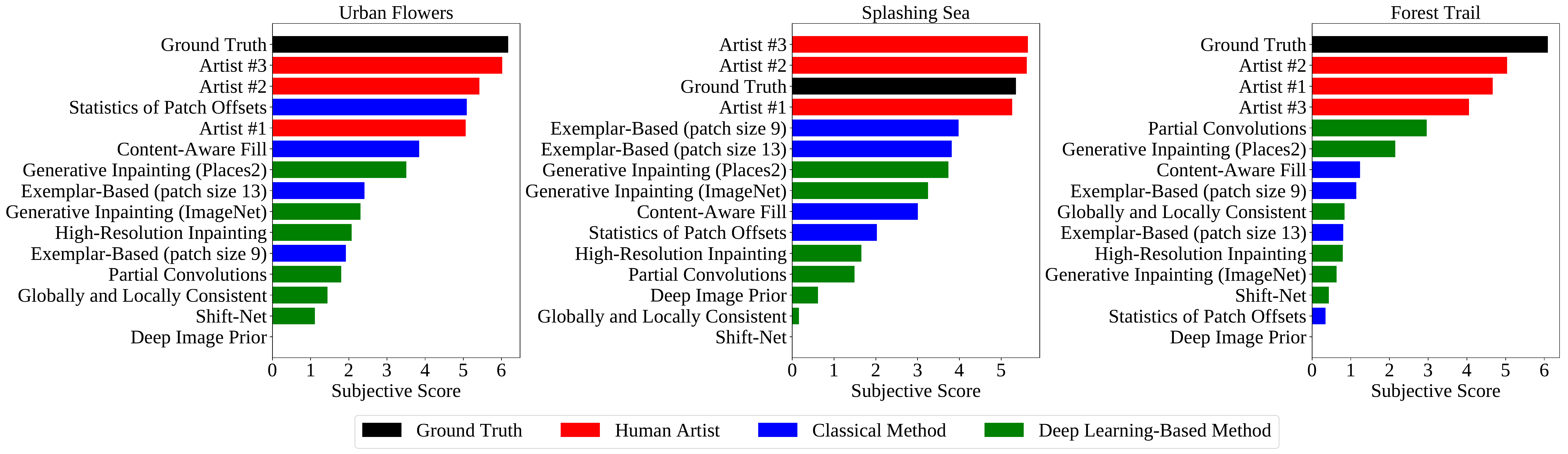}
\end{center}
   \caption{Results of the subjective study comparing images inpainted by human
            artists with images inpainted by conventional and
            deep-learning-based methods.}
   \label{fig:human-results-individual}
\end{figure*}

\subsection{Test method}

The subjective evaluation took place through the \url{http://subjectify.us}
platform.  Human observers were shown pairs of images and asked to pick from
each pair the one they found most realistic. Each image pair consisted of two
different inpainting results for the same picture (the set also contained the
original image). Also included were two verification questions that asked the
participants to compare the result of exemplar-based
inpainting~\cite{criminisi2004region} with the ground-truth image. The final
results excluded all responses from participants who failed to correctly answer
one or both verification questions. In total, 6,945 valid pairwise judgements
were collected from 215 participants.

The judgements were then used to fit a Bradley-Terry model
\cite{bradley1952rank}. The resulting subjective scores maximize likelihood
given the pairwise judgements.

\begin{figure}[t]
\begin{center}
   \setlength{\tabcolsep}{1pt}
   \begin{tabular}{cc}
      Artist \#1 & Statistics of Patch Offsets~\cite{he2012statistics} \\
      \includegraphics[width=0.49\linewidth]{media/human/city_flowers_0/fl_1.png}
      & \includegraphics[width=0.49\linewidth]{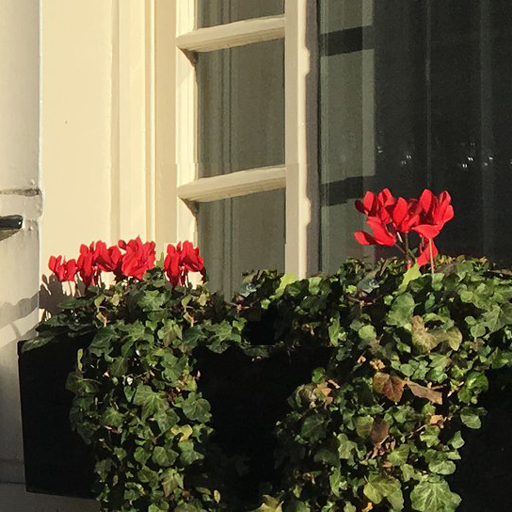}
   \end{tabular}
\end{center}
   \caption{Comparison of inpainting results from Artist~\#1 and statistics of
            patch offsets~\cite{he2012statistics} (preferred in the subjective
            comparison).}
   \label{fig:artist-1-patch-shift}
\end{figure}

\subsection{Results}

\noindent
\textbf{Algorithms vs. artists.}
Figure~\ref{fig:human-results-overall} shows the results for the three images
inpainted by the human artists. The artists outperformed all state-of-the-art
automatic algorithms, and out of the deep-learning-based methods, only
generative image inpainting~\cite{Yu_2018_CVPR} trained on the Places 2 data
set outperformed the classical inpainting methods.

The individual results for each of these three images appear in
Figure~\ref{fig:human-results-individual}. In only one case did an algorithm
beat an artist: statistics of patch offsets~\cite{he2012statistics} scored
higher than one artist on the ``Urban Flowers'' photo.
Figure~\ref{fig:artist-1-patch-shift} shows the respective inpainting results.
Additionally, for the ``Splashing Sea'' photo, two artists actually
``outperformed'' the original image: their results turned out to be more
realistic. This outcome illustrates that the ideal quality-estimation metric
should be a no-reference one.

\begin{figure}[t]
\begin{center}
   \includegraphics[width=\linewidth]{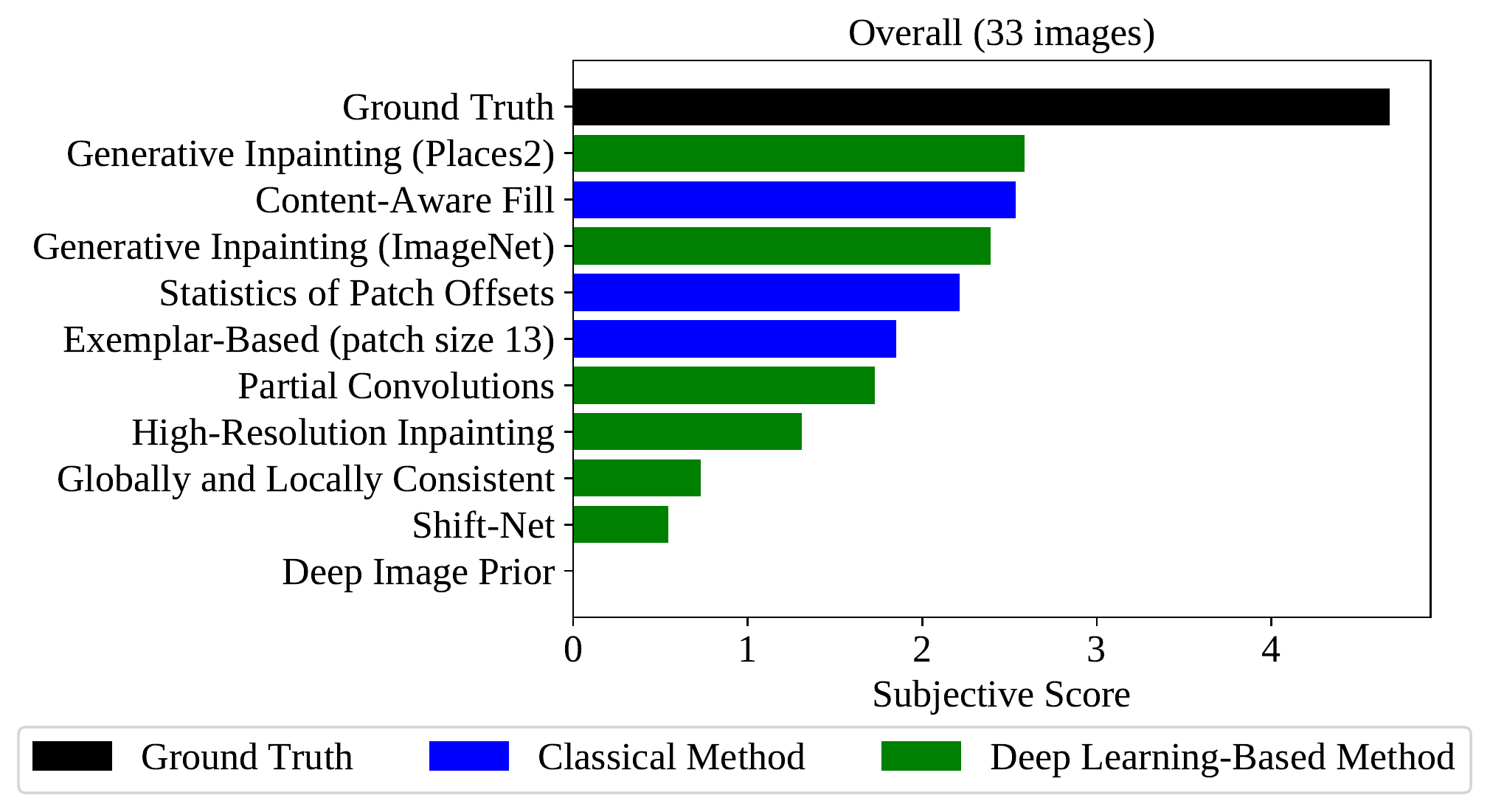}
\end{center}
   \caption{Subjective-comparison results for 33 images inpainted using
            automatic methods.}
   \label{fig:algorithm-results-overall}
\end{figure}

\bigskip
\noindent
\textbf{Algorithms vs. algorithms.}
We additionally performed a subjective comparison of various inpainting
algorithms among the entire 33-image test set, collecting 3,969 valid pairwise
judgements across 147 participants. The overall results appear in
Figure~\ref{fig:algorithm-results-overall}. They confirm our observations from
the first comparison: among the deep-learning-based approaches we evaluated,
generative image inpainting~\cite{Yu_2018_CVPR} seems to be the only one that
can outperform the classical methods.

\bigskip
\noindent
\textbf{Conclusion.} 
The subjective evaluation allows us to draw the following conclusions:
\begin{itemize}
   \item Manual inpainting by human artists remains the only viable way to
      achieve quality close to that of the original images.

   \item Automatic algorithms can approach manual-inpainting results only for
      certain images.

   \item Classical inpainting methods continue to maintain a strong position,
      even though a deep-learning-based algorithm took the lead.
\end{itemize}

\section{Objective inpainting-quality estimation}

Using the results we obtained from the subjective comparison, we evaluated
several approaches to objective inpainting-quality estimation. Using these
objective metrics, we estimated the inpainting quality of the images from our
test set and then compared them with the subjective results. For each of the 33
images, we applied every tested metric to every inpainting result (as well as
to the ground-truth image) and computed the Pearson and Spearman correlation
coefficients with the subjective result. The final value was an average of the
correlations over all 33 test images. 

Below is an overview of each metric we evaluated.

\subsection{Full-reference metrics}

Full-reference metrics take both the ground-truth (original) image and the
inpainting result as an input.

To construct a full-reference metric that encourages semantic similarity rather
than per-pixel similarity, as in~\cite{johnson2016perceptual}, we evaluated
metrics that compute the difference between the ground-truth and
inpainted-image feature maps produced by an image-classification neural
network. We selected five of the most popular deep architectures:
VGG~\cite{simonyan2014very} (16- and 19-layer deep variants),
ResNet\nobreakdash-V1\nobreakdash-50~\cite{He_2016_CVPR},
Inception\nobreakdash-V3~\cite{Szegedy_2016_CVPR},
Inception\nobreakdash-ResNet\nobreakdash-V2~\cite{szegedy2017inception} and
Xception~\cite{Chollet_2017_CVPR}. We used the models pretrained on the
ImageNet \cite{deng2009imagenet} data set. The mean squared error between the
feature maps was the metric result.

For VGG we tested the output from several deep layers (convolutional and
pooling layers from the last block) and found that the deepest layer has the
highest correlation with the subjective-study results. For each of the other
architectures, therefore, we tested only the deepest layer.

We additionally included the structural-similarity (SSIM)
index~\cite{wang2004image} as a full-reference metric. SSIM is widely used to
compare image quality, but it falls short when applied to inpainting-quality
estimation.

The best correlations among the full-reference metrics emerged when using the
last layer of the VGG\nobreakdash-16 model.

\subsection{No-reference metrics}

No-reference metrics take only the target (possibly inpainted) image as an
input, so they apply to a much wider range of problems.

We used a deep-learning approach to constructing no-reference metrics. We
picked several popular image-classification neural-network architectures and
trained them to differentiate ``clean'' (realistic, original) images without
any inpainting from partially inpainted images. The architectures included
VGG~\cite{simonyan2014very} (16- and 19-layer deep),
ResNet\nobreakdash-V1\nobreakdash-50~\cite{He_2016_CVPR},
ResNet\nobreakdash-V2\nobreakdash-50~\cite{he2016identity},
Inception\nobreakdash-V3~\cite{Szegedy_2016_CVPR},
Inception\nobreakdash-V4~\cite{szegedy2017inception} and
PNASNet\nobreakdash-Large~\cite{Liu_2018_ECCV_pnasnet}.

\bigskip
\noindent
\textbf{Data set.}
For training, we used clean and inpainted images based on the COCO
\cite{lin2014microsoft} data set. To create the inpainted images, we cropped
the input images to a square aspect ratio, resized them to $512\times512$
pixels and masked out a square of $180\times180$ pixels from the middle (the
same procedure we used when creating our subjective-evaluation data set). We
then inpainted the masked images using five inpainting
algorithms~\cite{criminisi2004region,iizuka2017globally,he2012statistics,Yan_2018_ECCV,Yu_2018_CVPR}
in eight total configurations. The total number of images in the data set
appears in Table~\ref{tab:training-data set-image-counts}.

\begin{table}
\begin{center}
   \begin{tabular}{rl}
      Exemplar-Based (patch size 9)~\cite{criminisi2004region} & 16777 \\
      Exemplar-Based (patch size 13)~\cite{criminisi2004region} & 16777 \\
      Globally and Locally Consistent~\cite{iizuka2017globally} & 13870 \\
      Statistics of Patch Offsets~\cite{he2012statistics} & 7955 \\
      Shift-Net~\cite{Yan_2018_ECCV} & 8928 \\
      Generative Inpainting (Places 2)~\cite{Yu_2018_CVPR} & 8928 \\
      Generative Inpainting (CelebA)~\cite{Yu_2018_CVPR} & 8928 \\
      Generative Inpainting (ImageNet)~\cite{Yu_2018_CVPR} & 8927 \\
      \hline
      Total inpainted & 91090 \\
      Total original (clean) & 81520
   \end{tabular}
\end{center}
   \caption{Total number of images in the training data set by inpainting
            algorithm.}
   \label{tab:training-data set-image-counts}
\end{table}

\begin{figure}[t]
\begin{center}
   \includegraphics[width=\linewidth]{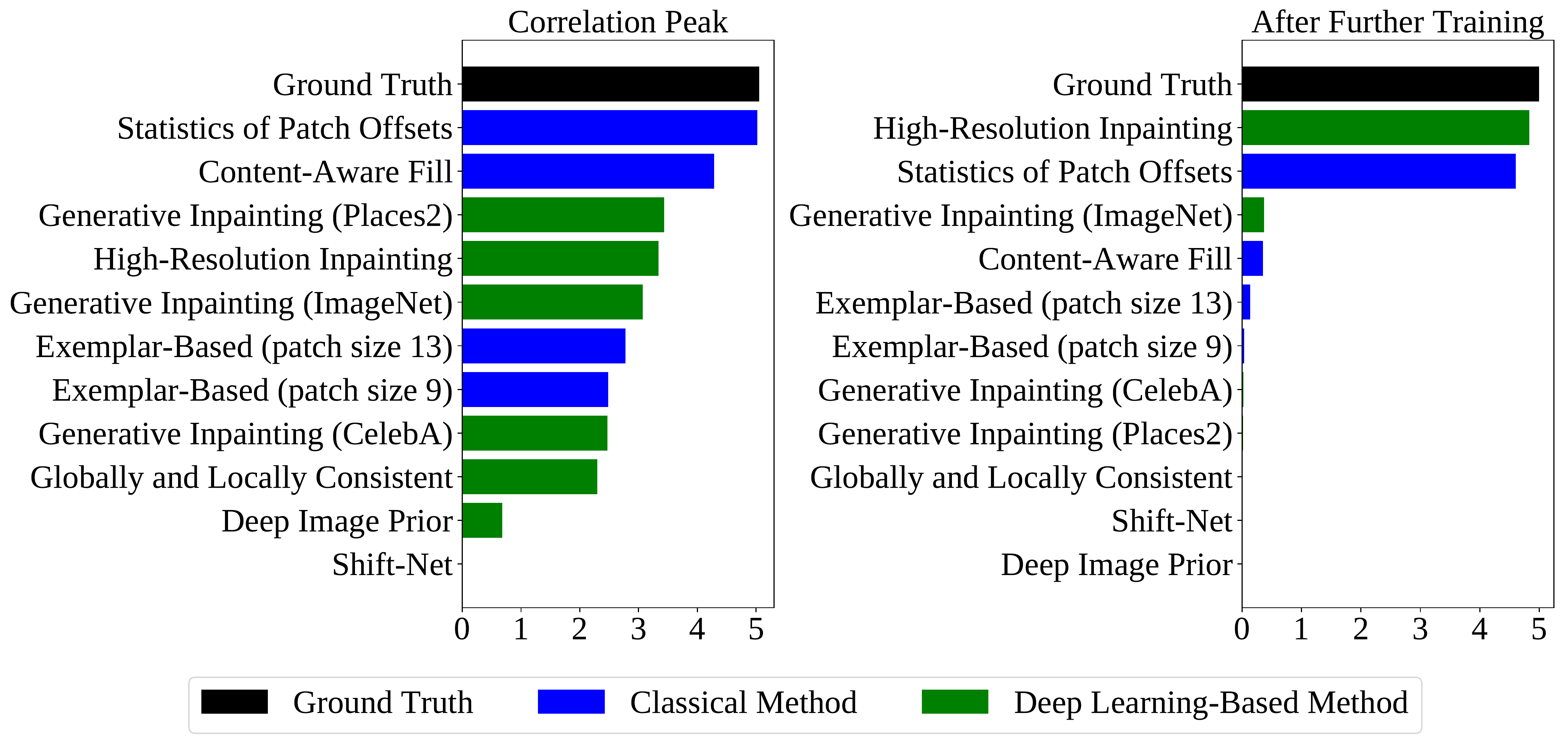}
\end{center}
   \caption{Inpainting quality estimated by VGG\nobreakdash-16 for one image at
            the peak Pearson correlation and after further training. The score
            distribution at the correlation peak is similar to a
            subjective-score distribution; after further training, however, the
            network starts to heavily underscore most inpainting algorithms.}
   \label{fig:overfitting}
\end{figure}

\begin{figure*}
\begin{center}
   \includegraphics[width=\linewidth]{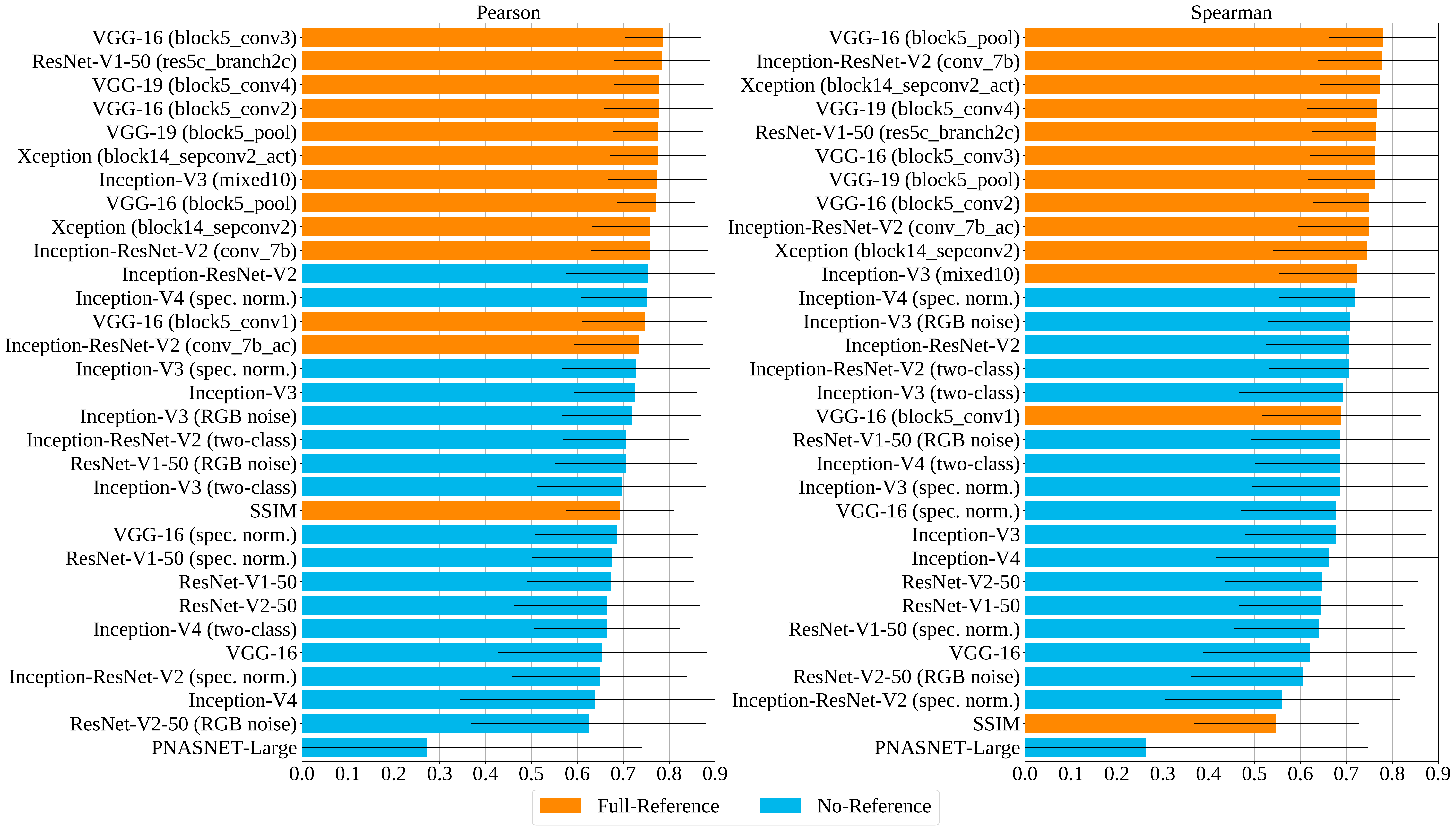}
\end{center}
   \caption{Mean Pearson and Spearman correlations between objective
            inpainting-quality metrics and subjective human comparisons
            (including ground-truth scores). The error bars show the standard
            deviations.}
   \label{fig:correlations}
\end{figure*}

\begin{figure*}
\begin{center}
   \includegraphics[width=\linewidth]{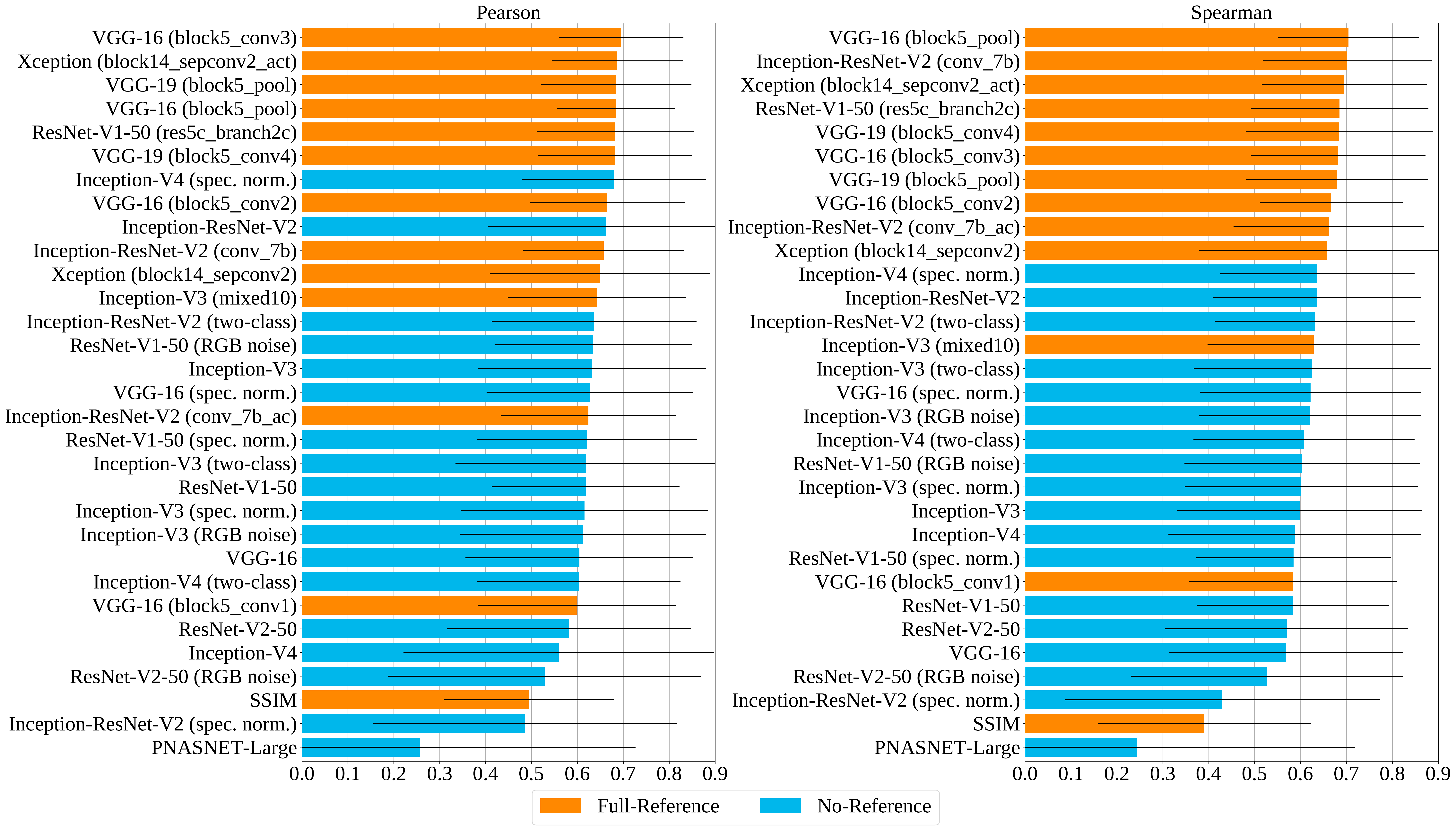}
\end{center}
   \caption{Mean Pearson and Spearman correlations between objective
            inpainting-quality metrics and subjective human comparisons
            (excluding ground-truth scores). The error bars show the standard
            deviations.}
   \label{fig:correlations-no-gt}
\end{figure*}

\bigskip
\noindent
\textbf{Training.}
The network architectures take a square image as an input and output the
score---a single number where 0 means the image contains inpainted regions and
1 means the image is ``clean.'' The loss function was mean squared error. Some
network architectures were additionally trained to output the predicted class
using one-hot encoding (similar to binary classification); the loss function
for this case was softmax cross-entropy.

The network architectures were identical to the ones used for image
classification, with one difference: we altered the number of outputs from the
last fully connected layer. This change allowed us to initialize the weights of
all previous layers from the models pretrained on ImageNet, greatly improving
the results compared with training from random initialization.

For some experiments we additionally tried using the RGB noise features
described in~\cite{Zhou_2018_CVPR} and the spectral weight normalization
described in~\cite{miyato2018spectral}.

In addition to the typical validation on part of the data set, we also
monitored correlation of network predictions with the subjective scores
collected in Section~\ref{sec:subjective-evaluation}. We used the networks to
estimate the inpainting quality of the 33-image test set, then computed
correlations with subjective results in the same way as the final comparison.
The training of each network was stopped once the correlation of the network
predictions with the subjective scores peaked and started to decrease (possibly
because the networks were overfitting to the inpainting results of the
algorithms we used to create the training data set).
Figure~\ref{fig:overfitting} compares the network scores for one test image at
the correlation peak and after further training.

\subsection{Results}

We evaluated several objective quality-estimation approaches, both full
reference and no reference as well as both classical and deep-learning based.
Figure~\ref{fig:correlations} shows the overall results. We additionally did a
comparison that excluded ground-truth scores from the correlation, because the
full-reference approaches always yield the highest score when comparing the
ground-truth image with itself. The overall results for that comparison appear
in Figure~\ref{fig:correlations-no-gt}.

As Figures~\ref{fig:correlations} and~\ref{fig:correlations-no-gt} show, the
no-reference methods achieve slightly weaker correlation with the
subjective-evaluation responses than do the best full-reference methods. But
the results of most no-reference methods are still considerably better than
those of the full-reference SSIM. The best correlations among the no-reference
methods came from the Inception\nobreakdash-V4 model trained to output one
score, with spectral weight normalization. 

It is important to emphasize that we did \textit{not} train the networks to maximize
correlation with human responses or with the subjective-evaluation data set in
general. We trained them to distinguish ``clean'' images from inpainted images,
yet their output showed good correlation with human responses.

\section{Conclusion}

We have proposed a number of perceptually motivated no-reference and
full-reference objective metrics for evaluating image-inpainting quality. We
evaluated the metrics by correlating them with human responses from a
subjective comparison of state-of-the-art image-inpainting algorithms.

The results of the subjective comparison indicate that although a
deep-learning-based approach to image inpainting holds the lead, classical
algorithms remain among the best in the field. 

The highest mean Pearson correlation with the human responses from the
subjective study was achieved by the block5\_conv3 layer of the
VGG\nobreakdash-16 model trained on ImageNet (full reference) and by the
Inception\nobreakdash-V4 model with spectral normalization (no reference). 

We achieved good correlation with the subjective\-/comparison results without
specifically training our proposed objective quality-evaluation metrics on the
subjective-comparison response data set. 

{\small
\bibliographystyle{ieee}
\bibliography{references}

\begin{thebibliography}{10}\itemsep=-1pt

\bibitem{Bappy_2017_ICCV}
J.~H. Bappy, A.~K. Roy-Chowdhury, J.~Bunk, L.~Nataraj, and B.~S. Manjunath.
\newblock Exploiting spatial structure for localizing manipulated image
  regions.
\newblock In {\em The IEEE International Conference on Computer Vision (ICCV)},
  Oct 2017.

\bibitem{bradley1952rank}
R.~A. Bradley and M.~E. Terry.
\newblock Rank analysis of incomplete block designs: I. the method of paired
  comparisons.
\newblock {\em Biometrika}, 39(3/4):324--345, 1952.

\bibitem{Chollet_2017_CVPR}
F.~Chollet.
\newblock Xception: Deep learning with depthwise separable convolutions.
\newblock In {\em The IEEE Conference on Computer Vision and Pattern
  Recognition (CVPR)}, July 2017.

\bibitem{criminisi2004region}
A.~Criminisi, P.~P{\'e}rez, and K.~Toyama.
\newblock Region filling and object removal by exemplar-based image inpainting.
\newblock {\em IEEE Transactions on Image Processing}, 13(9):1200--1212, 2004.

\bibitem{deng2009imagenet}
J.~Deng, W.~Dong, R.~Socher, L.-J. Li, K.~Li, and L.~Fei-Fei.
\newblock Imagenet: A large-scale hierarchical image database.
\newblock In {\em 2009 IEEE conference on computer vision and pattern
  recognition}, pages 248--255. Ieee, 2009.

\bibitem{he2012statistics}
K.~He and J.~Sun.
\newblock Statistics of patch offsets for image completion.
\newblock In {\em European Conference on Computer Vision}, pages 16--29.
  Springer, 2012.

\bibitem{He_2016_CVPR}
K.~He, X.~Zhang, S.~Ren, and J.~Sun.
\newblock Deep residual learning for image recognition.
\newblock In {\em The IEEE Conference on Computer Vision and Pattern
  Recognition (CVPR)}, June 2016.

\bibitem{he2016identity}
K.~He, X.~Zhang, S.~Ren, and J.~Sun.
\newblock Identity mappings in deep residual networks.
\newblock In {\em European conference on computer vision}, pages 630--645.
  Springer, 2016.

\bibitem{iizuka2017globally}
S.~Iizuka, E.~Simo-Serra, and H.~Ishikawa.
\newblock Globally and locally consistent image completion.
\newblock {\em ACM Transactions on Graphics (ToG)}, 36(4):107, 2017.

\bibitem{johnson2016perceptual}
J.~Johnson, A.~Alahi, and L.~Fei-Fei.
\newblock Perceptual losses for real-time style transfer and super-resolution.
\newblock In {\em European conference on computer vision}, pages 694--711.
  Springer, 2016.

\bibitem{li2018context}
H.~Li, G.~Li, L.~Lin, H.~Yu, and Y.~Yu.
\newblock Context-aware semantic inpainting.
\newblock {\em IEEE Transactions on Cybernetics}, 2018.

\bibitem{li2017image}
H.~Li, W.~Luo, X.~Qiu, and J.~Huang.
\newblock Image forgery localization via integrating tampering possibility
  maps.
\newblock {\em IEEE Transactions on Information Forensics and Security},
  12(5):1240--1252, 2017.

\bibitem{lin2014microsoft}
T.-Y. Lin, M.~Maire, S.~Belongie, J.~Hays, P.~Perona, D.~Ramanan,
  P.~Doll{\'a}r, and C.~L. Zitnick.
\newblock Microsoft coco: Common objects in context.
\newblock In {\em European conference on computer vision}, pages 740--755.
  Springer, 2014.

\bibitem{Liu_2018_ECCV_pnasnet}
C.~Liu, B.~Zoph, M.~Neumann, J.~Shlens, W.~Hua, L.-J. Li, L.~Fei-Fei,
  A.~Yuille, J.~Huang, and K.~Murphy.
\newblock Progressive neural architecture search.
\newblock In {\em The European Conference on Computer Vision (ECCV)}, September
  2018.

\bibitem{Liu_2018_ECCV}
G.~Liu, F.~A. Reda, K.~J. Shih, T.-C. Wang, A.~Tao, and B.~Catanzaro.
\newblock Image inpainting for irregular holes using partial convolutions.
\newblock In {\em The European Conference on Computer Vision (ECCV)}, September
  2018.

\bibitem{liu2017semantically}
P.~Liu, X.~Qi, P.~He, Y.~Li, M.~R. Lyu, and I.~King.
\newblock Semantically consistent image completion with fine-grained details.
\newblock {\em arXiv preprint arXiv:1711.09345}, 2017.

\bibitem{miyato2018spectral}
T.~Miyato, T.~Kataoka, M.~Koyama, and Y.~Yoshida.
\newblock Spectral normalization for generative adversarial networks.
\newblock {\em arXiv preprint arXiv:1802.05957}, 2018.

\bibitem{Pathak_2016_CVPR}
D.~Pathak, P.~Krahenbuhl, J.~Donahue, T.~Darrell, and A.~A. Efros.
\newblock Context encoders: Feature learning by inpainting.
\newblock In {\em The IEEE Conference on Computer Vision and Pattern
  Recognition (CVPR)}, June 2016.

\bibitem{pun2015image}
C.-M. Pun, X.-C. Yuan, and X.-L. Bi.
\newblock Image forgery detection using adaptive oversegmentation and feature
  point matching.
\newblock {\em IEEE Transactions on Information Forensics and Security},
  10(8):1705--1716, 2015.

\bibitem{salloum2018image}
R.~Salloum, Y.~Ren, and C.-C.~J. Kuo.
\newblock Image splicing localization using a multi-task fully convolutional
  network (mfcn).
\newblock {\em Journal of Visual Communication and Image Representation},
  51:201--209, 2018.

\bibitem{simonyan2014very}
K.~Simonyan and A.~Zisserman.
\newblock Very deep convolutional networks for large-scale image recognition.
\newblock {\em arXiv preprint arXiv:1409.1556}, 2014.

\bibitem{song2017image}
Y.~Song, C.~Yang, Z.~Lin, H.~Li, Q.~Huang, and C.~J. Kuo.
\newblock Image inpainting using multi-scale feature image translation.
\newblock {\em arXiv preprint arXiv:1711.08590}, 2, 2017.

\bibitem{szegedy2017inception}
C.~Szegedy, S.~Ioffe, V.~Vanhoucke, and A.~A. Alemi.
\newblock Inception-v4, inception-resnet and the impact of residual connections
  on learning.
\newblock In {\em Thirty-First AAAI Conference on Artificial Intelligence},
  2017.

\bibitem{Szegedy_2016_CVPR}
C.~Szegedy, V.~Vanhoucke, S.~Ioffe, J.~Shlens, and Z.~Wojna.
\newblock Rethinking the inception architecture for computer vision.
\newblock In {\em The IEEE Conference on Computer Vision and Pattern
  Recognition (CVPR)}, June 2016.

\bibitem{telea2004image}
A.~Telea.
\newblock An image inpainting technique based on the fast marching method.
\newblock {\em Journal of Graphics Tools}, 9(1):23--34, 2004.

\bibitem{Ulyanov_2018_CVPR}
D.~Ulyanov, A.~Vedaldi, and V.~Lempitsky.
\newblock Deep image prior.
\newblock In {\em The IEEE Conference on Computer Vision and Pattern
  Recognition (CVPR)}, June 2018.

\bibitem{wang2004image}
Z.~Wang, A.~C. Bovik, H.~R. Sheikh, E.~P. Simoncelli, et~al.
\newblock Image quality assessment: from error visibility to structural
  similarity.
\newblock {\em IEEE transactions on image processing}, 13(4):600--612, 2004.

\bibitem{Yan_2018_ECCV}
Z.~Yan, X.~Li, M.~Li, W.~Zuo, and S.~Shan.
\newblock Shift-net: Image inpainting via deep feature rearrangement.
\newblock In {\em The European Conference on Computer Vision (ECCV)}, September
  2018.

\bibitem{Yang_2017_CVPR}
C.~Yang, X.~Lu, Z.~Lin, E.~Shechtman, O.~Wang, and H.~Li.
\newblock High-resolution image inpainting using multi-scale neural patch
  synthesis.
\newblock In {\em The IEEE Conference on Computer Vision and Pattern
  Recognition (CVPR)}, July 2017.

\bibitem{yu2018free}
J.~Yu, Z.~Lin, J.~Yang, X.~Shen, X.~Lu, and T.~S. Huang.
\newblock Free-form image inpainting with gated convolution.
\newblock {\em arXiv preprint arXiv:1806.03589}, 2018.

\bibitem{Yu_2018_CVPR}
J.~Yu, Z.~Lin, J.~Yang, X.~Shen, X.~Lu, and T.~S. Huang.
\newblock Generative image inpainting with contextual attention.
\newblock In {\em The IEEE Conference on Computer Vision and Pattern
  Recognition (CVPR)}, June 2018.

\bibitem{Zhou_2018_CVPR}
P.~Zhou, X.~Han, V.~I. Morariu, and L.~S. Davis.
\newblock Learning rich features for image manipulation detection.
\newblock In {\em The IEEE Conference on Computer Vision and Pattern
  Recognition (CVPR)}, June 2018.

\bibitem{zhu2018deep}
X.~Zhu, Y.~Qian, X.~Zhao, B.~Sun, and Y.~Sun.
\newblock A deep learning approach to patch-based image inpainting forensics.
\newblock {\em Signal Processing: Image Communication}, 67:90--99, 2018.

\end{thebibliography}
}

\end{document}